\pdfoutput=1

\documentclass{article}

\usepackage[preprint,nonatbib]{neurips_2024}

\usepackage[utf8]{inputenc} 
\usepackage[T1]{fontenc}    
\usepackage{hyperref}       
\usepackage{url}            
\usepackage{booktabs}       
\usepackage{amsfonts}       
\usepackage{nicefrac}       
\usepackage{microtype}      
\usepackage{xcolor}         
\usepackage{amsmath}
\usepackage{authblk}   
\usepackage{float}
\usepackage{graphicx}
\usepackage{subcaption}
\usepackage[backend=biber, style=numeric, sorting=none]{biblatex}
\title{ClinicalGPT-R1: Pushing reasoning capability of generalist disease diagnosis with large language model}

\author[1]{Wuyang Lan\textsuperscript{\#}}
\author[1]{Wenzheng Wang\textsuperscript{\#}}
\author[1]{Changwei Ji} 
\author[1]{Guoxing Yang}
\author[2]{Yongbo Zhang}
\author[2]{Xiaohong Liu}
\author[3]{Luonan Chen}
\author[4]{Shengge Li}
\author[2]{Song Wu}
\author[1]{Guangyu Wang\textsuperscript{*}}

\affil[1]{State Key Laboratory of Networking and Switching Technology \protect\\ 
          Beijing University of Posts and Telecommunications}
\affil[2]{South China Hospital, Medical School, Shenzhen University}
\affil[3]{School of Mathematical Sciences and School of AI, Shanghai Jiao Tong University}
\affil[4]{China Mobile Group China Mobile Internet Co.,Ltd.}

\begin{document}

\maketitle

\begin{abstract}

Recent advances in reasoning with large language models (LLMs)has shown remarkable reasoning capabilities in domains such as mathematics and coding, yet their application to clinical diagnosis remains underexplored. Here, we introduce ClinicalGPT-R1, a reasoning-enhanced generalist large language model for disease diagnosis. Trained on a dataset of 20,000 real-world clinical records, ClinicalGPT-R1 leverages diverse training strategies to enhance diagnostic reasoning. 
To benchmark performance, we curated MedBench-Hard, a challenging dataset spanning seven major medical specialties and representative diseases. Experimental results demonstrate that ClinicalGPT-R1 outperforms GPT-4o in Chinese diagnostic tasks and achieves comparable performance to GPT-4o in English settings. This comparative study effectively validates the superior performance of ClinicalGPT-R1 in disease diagnosis tasks. Resources are available at \url{https://github.com/medfound/medfound}.

\end{abstract}

\renewcommand{\thefootnote}{}
\footnotetext{\#These authors contributed equally to this work.}
\footnotetext{* Corresponding Author: \href{mailto:guangyu.wang24@gmail.com}{guangyu.wang24@gmail.com}}
\renewcommand{\thefootnote}{\arabic{footnote}}

\section{Introduction}
Recent advances in large language models (LLMs), such as OpenAI-o1\cite{guan2024deliberative} and DeepSeek-R1\cite{8}, have demonstrated strong reasoning abilities across domains including mathematics and programming. \cite{zhong2024evaluation}\cite{guo2025deepseek}. These developments highlight the value of scaling techniques like extended Chain-of-Thought (CoT) prompting and reinforcement learning to enhance model reasoning\cite{4}\cite{5}. Motivated by these general-purpose breakthroughs, there has been a growing interest to explore LLMs in domain-specific settings, especially in healthcare\cite{10}\cite{griot2025large}. In clinical contexts, robust long-form reasoning is essential for accurate diagnosis and treatment.\cite{sayed2025undergraduate}. Therefore, advancing and evaluating reasoning-augmented LLMs for medical applications is of critical importance.

A key challenge in medical reasoning with LLMs lies in validating their reasoning processes\cite{6}. In domains like mathematics or programming, intermediate steps can be quantitatively assessed and leveraged for supervised training or reinforcement learning\cite{9,11}. For instance, code compilation or step-wise annotation provides clear feedback signals. In contrast, clinical reasoning—especially in real-world diagnostic scenarios—often lacks such structured, verifiable steps\cite{lightman2023let}. While models like HuatuoGPT-O1 have utilized multiple-choice datasets (e.g., MedQA) to simulate clinical reasoning\cite{6}, these fall short of capturing the complexity of authentic diagnostic workflows. This highlights the need for datasets specifically tailored to real-world clinical reasoning.

The effectiveness of medical reasoning with LLMs is shaped by multiple factors, including training strategies, the quality of synthetic data, and languages. In the general domain, Deepseek-R1\cite{8} significantly enhances the reasoning capabilities of large models through a purely reinforcement learning approach. In medical reasoning tasks, Marco-O1 \cite{zhao2411marco} uses Monte Carlo Tree Search (MCTS) to construct process-supervised datasets, optimizing model training by using confidence scores as the value for each reasoning step. Zhou et al. \cite{zhou2024jiuzhang3} improve the mathematical reasoning ability of small models by synthesizing mathematical reasoning data using GPT-4, which is then used to train a smaller LLM to efficiently generate high-quality In healthcare, efforts have been made to synthesize dialogue datasets from real-world electronic health records\cite{fan2024ai}, and recent work has explored generating long-chain reasoning text tailored to clinical scenarios\cite{6,huang2025o1}. Yet, a systematic investigation into how these factors influence diagnostic reasoning in real-world medical contexts remains lacking.

In this study, we introduce ClinicalGPT-R1, a large language model specifically designed for clinical reasoning tasks in Chinese and English medical settings. Firstly, leveraging two state-of-the-art LLMs as data synthesizers, we construct a clinical diagnostic dataset grounded in real clinical records, encompassing clinical reasoning tasks. Through a two-stage training approach combining supervised fine-tuning (SFT) and reinforcement learning (RL), the model's reasoning capacity is improved to better handle clinical tasks in real-world scenarios. To benchmark performance, we introduce MedBench-Hard, a diverse and challenging dataset spanning seven major medical departments. Experimental results show that ClinicalGPT-R1 significantly outperforms GPT-4o in Chinese tasks, while achieving a comparable performance to GPT-4o in English tasks. This demonstrates the effectiveness of our approach and its strong practical capabilities in generalist disease diagnosis.

\section{Methods}
\subsection{Medical Data}
\subsubsection{Data source}
To enhance the medical capabilities of our model, we dedicate significant efforts to collecting large-scale, high-quality, and diverse medical data for training. We employ a sophisticated data classification and filtering process to improve data quality and ensure comprehensive coverage of medical domains. Furthermore, synthetic data has been shown to be effective for model training\cite{12}, and thus we generate extensive synthetic data to complement medical data. Specifically, we utilize various advanced data synthesis techniques and design tailored synthesis strategies for different data sources, resulting in data of substantial scale, diverse formats, high educational value, and full coverage of medical scenarios.

Our training corpus is derived from two primary sources. The first dataset originates from MedDX-FT\cite{1}, which includes extensive medical records from multiple departments and the corresponding diagnostic reasoning cases to generate complex reasoning data. The second source comprises real-world medical data, including anonymized electronic health records (EHRs). 

\subsubsection{Synthetic Data}
\label{subsec:synthetic_data}

\begin{figure}[htbp]
    \centering
    \includegraphics[width=0.6\textwidth]{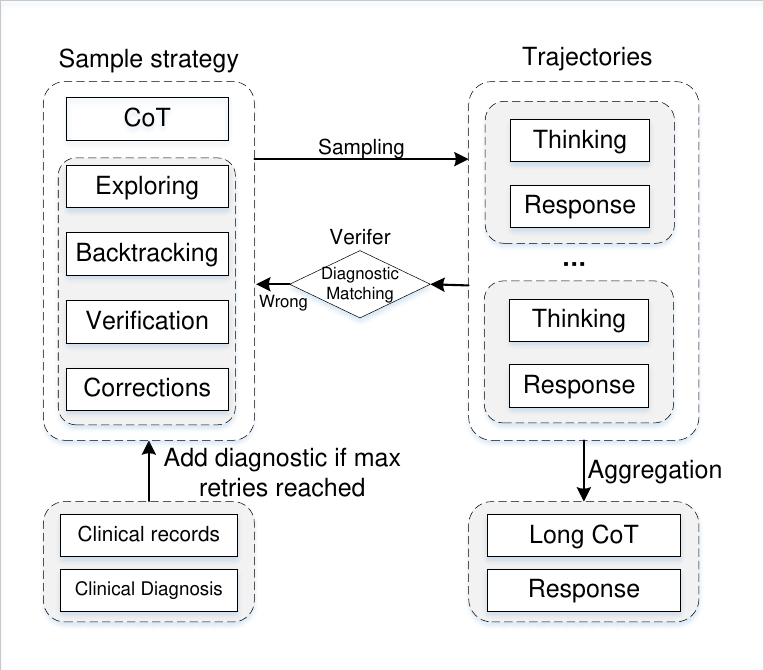} 
    \caption{The pipeline of synthesized data} 
    \label{fig:0} 
\end{figure}

The use of synthetic data has become a cornerstone in training LLMs, particularly for domains with limited annotated resources\cite{zhao2411marco}. To improve the quality of synthetic medical data, we developed a generation pipeline with various strategies grounded in real-world medical records and diagnostic outcomes\ref{fig:0}. 
In the design of data construction processes, we emphasize the importance of the reasoning process and long-chain thinking capabilities. Specifically, we first leverage state-of-the-art LLMs combined with a long-chain reasoning prompt strategy to generate both steps and final results\cite{savage2024diagnostic}. If the reasoning process leads to a correct outcome, it will be used directly. If the correct result cannot be generated, we apply four distinct strategies \cite{7}: Exploring New Paths, Backtracking, Verification, and Corrections. When the maximum iteration count is reached, the search process is restarted. Each data point is attempted up 3 times; if all attempts fail, we will make one final attempt. In this case, we directly provide the reference answer in the prompt and outline the CoT reasoning path, guiding the model to generate the final inference process.

In order to enhance the reasoning process, we employ a dynamic approach that reformats the reasoning trajectory into a coherent, natural language-based process, referred to as long CoT. This reformatting avoids rigid structures, incorporating smooth transition words (e.g., "hmm," "also," "wait") to streamline reasoning and reduce token usage. Subsequently, the model generates a formal response, referred to as the long response, based on the conclusion derived from the long CoT to address the given question.

\subsubsection{Evaluation benchmark: MedBench-Hard}
To comprehensively assess disease diagnostic capabilities, we have constructed a disease diagnosis benchmark, MedBench-Hard. Specifically, we collected clinical diagnostic data from seven different departments, with 500 cases per department, totaling 3,500 test samples. The departments included Respiratory, Gastroenterology, Urology, Cardiology, Immunology, Neurology, and Endocrinology. To ensure a diverse representation of diseases across each department and avoid complete duplication of conditions within a single department, we employ a stratified sampling approach. Specifically, we performed stratified sampling based on International Classification of Diseases(ICD)-10 disease codes, selecting diseases from various levels to maximize coverage of the different types of disease within each department, including rare diseases. 

\subsection{Post-train}

\subsubsection{Supervised Fine-Tuning}
We meticulously curated a medical instruction fine-tuning dataset through the pipeline in \ref{subsec:synthetic_data}.
The supervised fine-tuning (SFT) data are made up mainly of three key components: the question, the thinking, and the final response. The "question" section includes the patient's medical history and anonymized clinical records; the "thinking" refers to the reasoning sequence, which represents the model's explicit cognitive steps; and the "final response" is the outcome of this reasoning, offering a comprehensive symptom analysis and diagnostic conclusion.

\subsubsection{Reinforcement Learning}
In this stage, we further enhance the long-term reasoning capabilities of the model through RL. Although the LLM has already learned successful reasoning trajectories during the SFT phase, which were derived through search, these trajectories may not be optimal. The On-policy learning in the RL phase aims to optimize the model for better performance in long CoT reasoning.

Rewards play a vital role in steering the target of RL training. 
We employ a result-based reward design method. This approach rewards data points that exhibit a clear reasoning process and provide correct responses, while penalizing those that directly present diagnostic results without a reasoning process, even if the results are correct.
Following [\cite{6}, \cite{riedmiller2018learning},  \cite{trott2023solving}], the reward is assigned as

\begin{equation}
r'(x, \hat{y}, y^*) =
\begin{cases}
1 & \text{if } \text{verifier}(\hat{y}, y^*) = \text{True} \\
0.1 & \text{if } \text{verifier}(\hat{y}, y^*) = \text{False} \\
0 & \text{if } \hat{y} = \text{null}
\end{cases}
\end{equation}

Correct answers are awarded a reward of 1, incorrect answers receive 0.1, and responses that demonstrate a lack of "think-before-answering" behavior are assigned a reward of 0. We transform medical inquiries into verifiable questions, obtaining quantifiable feedback through a large-scale model verifier. 

We transform medical inquiries into verifiable questions, obtaining quantifiable feedback through a large-scale model verifier. The Rule-Based Reward Model (RM) can indeed cover a broad range of domains and applications. Here, we use the Proximal Policy Optimization (PPO) algorithm \cite{3} as an RL technique to further optimize the decision-making process within the model.

\section{Evaluations}
\subsection{Overview}
Our evaluation is conducted in real-world clinical settings, evaluating the performance of seven departments to generate accurate results, in order to comprehensively assess the impact of various factors on the reasoning and comprehension capabilities of the model.
We conducted a series of comparative experiments in ClinicalGPT-R1 to explore the effects of different training strategies, languages, and the quality of synthetic data. Furthermore, we benchmark the ClinicalGPT-R1 model against Qwen2.5-7B-Instruct\cite{2} and GPT-4o to assess its competitiveness in medical diagnostic tasks.

\subsection{Implemented details}
We use Qwen-2.5-7B-instruct as the base model and fine-tune it for 3 epochs, employing a cosine decay learning rate schedule that starts at \(5 \times 10^{-6}\) and gradually decreases to \(5 \times 10^{-6}\) in the SFT stage. During training, each individual sequence is formed by packing multiple samples. In addition, we implement a sample masking strategy to ensure that these examples remain isolated and invisible to each other. In the RL stage, the learning rate of \( 5 \times 10^{-7} \), the batch size of 16, and a \( \beta \) value set to 0.03. The PPO parameters are configured as follows: 3 epochs, a discount factor of 1.0, a value coefficient of 1.0, and a clip range of 0.2. 

\subsection{Experiment results}
\begin{figure}[htbp]
    \centering
    \includegraphics[width=1\textwidth]{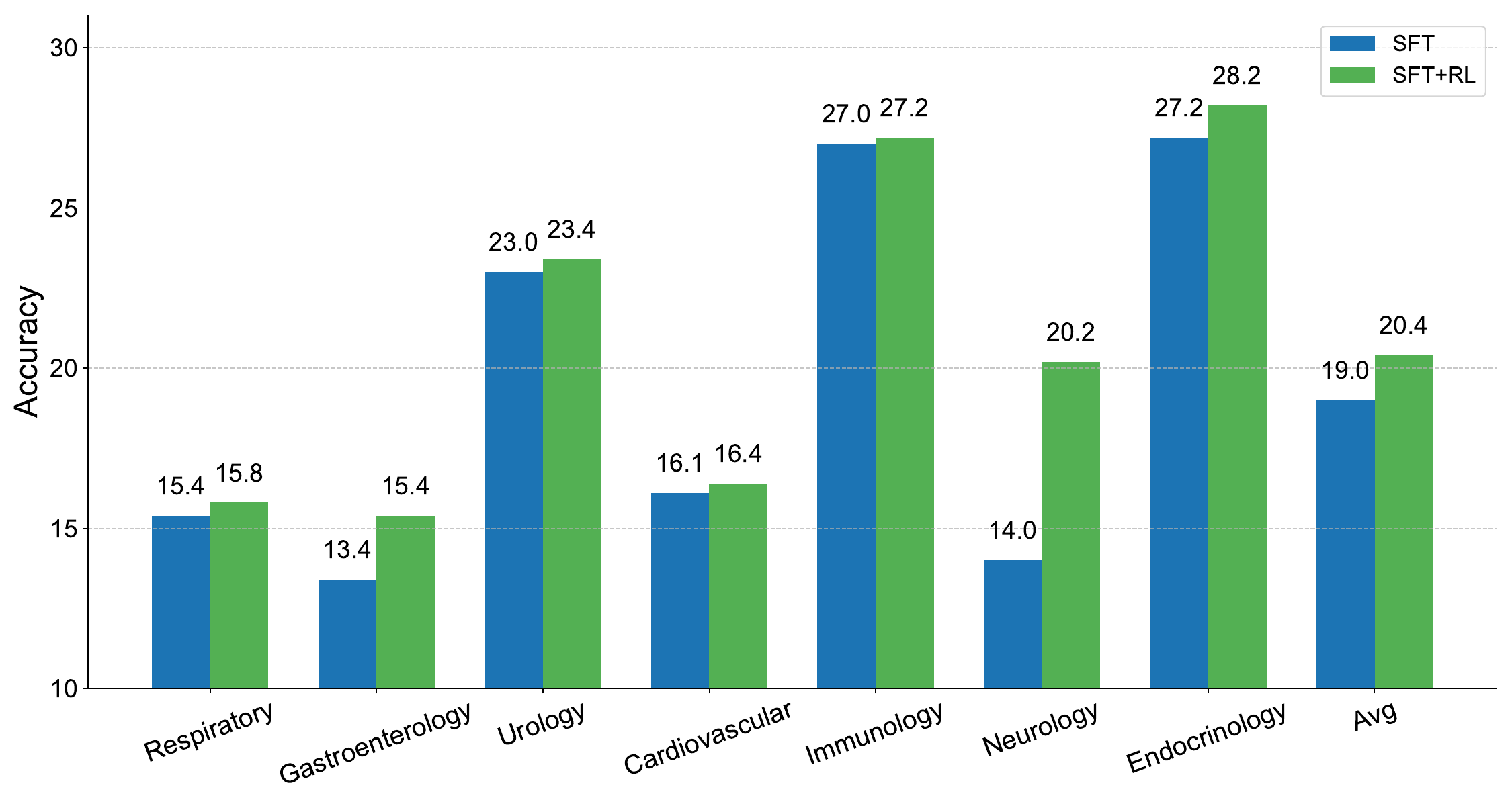} 
    \caption{Comparison of ClinicalGPT-R1 with different training methods} 
    \label{fig:1} 
\end{figure}

To validate the effectiveness of the two-stage training method, we compared the performance of the model trained solely with SFT with that of the model trained with an additional RL-based reinforcement for enhanced reasoning, following the SFT phase. Firstly, SFT enhances the model's reasoning ability on the basis of the base model, aiming to enable the model to learn thinking before solving medical problems. Then, the model's decision-making process is further optimized through reinforcement learning. The model is trained in a simulated medical decision-making environment to improve its reasoning and decision-making capabilities. This stage of training aims to enable the model to engage in deep reasoning before making more informed and rational decisions in real-world medical scenarios. As shown in Fig. ~\ref{fig:1}, the model trained using a combination of SFT and RL demonstrates superior performance in reasoning and diagnostic tasks compared to the model trained solely with SFT.

\begin{figure}[htbp]
    \centering
    \includegraphics[width=1\textwidth]{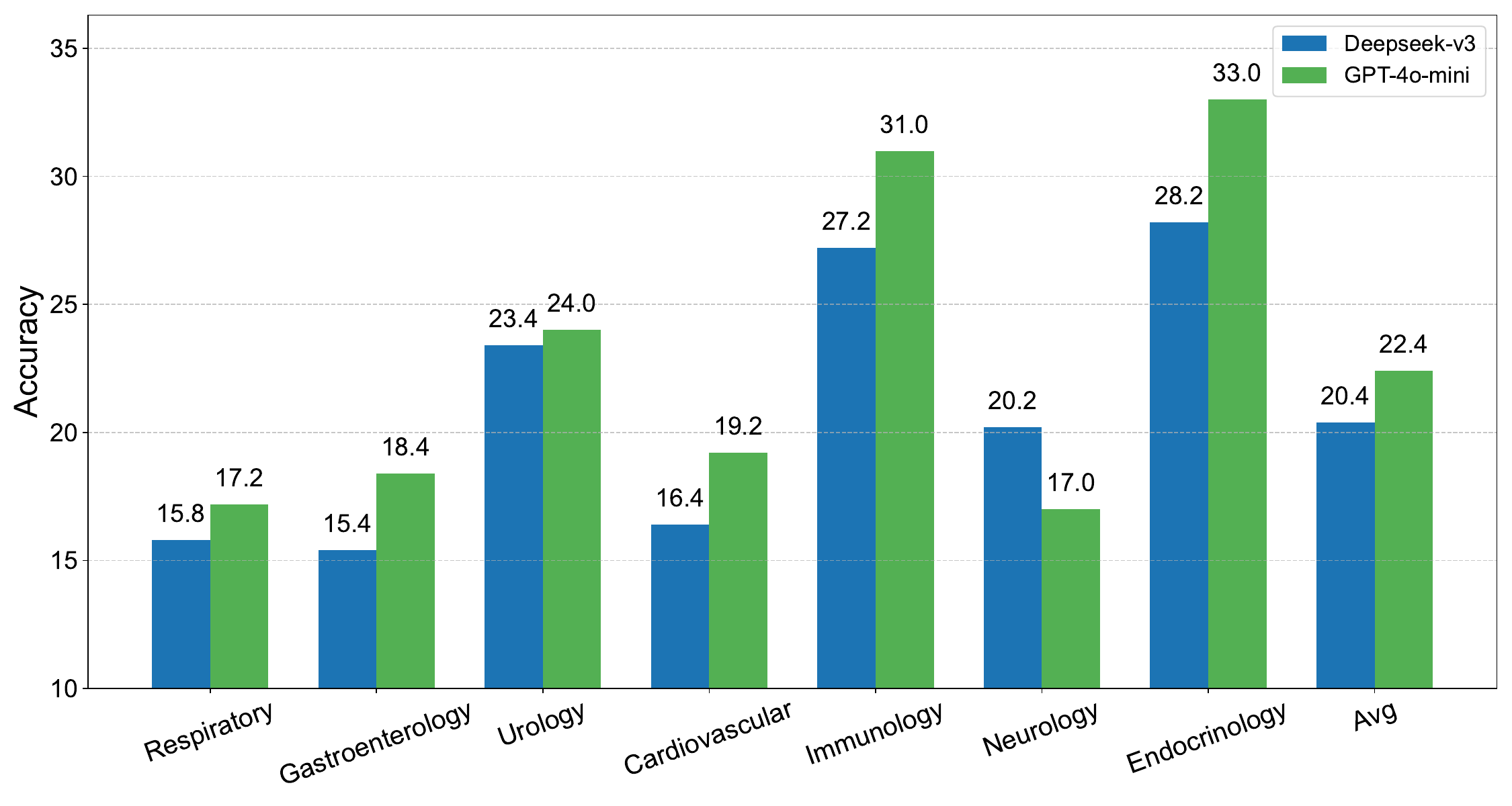} 
    \caption{Comparison of ClinicalGPT-R1 with different training data sources} 
    \label{fig:2} 
\end{figure}

The quality of synthetic data is crucial for the learning performance of LLMs. In this experiment, we selected two powerful models: GPT-4o-mini and Deepseek-v3-0324, as data generators. We then used the complex reasoning data synthesized by these models as input for the SFT and RL training of the LLM. We conducted comparative experiments on different data sources (Fig. ~\ref{fig:2}). As shown in the data in the figure, the model trained with the data generated by GPT-4o-mini performed better in the medical diagnosis test than the model trained with the data generated by Deepseek-v3-0324.

\begin{figure}[htbp]
    \centering
    \includegraphics[width=1\textwidth]{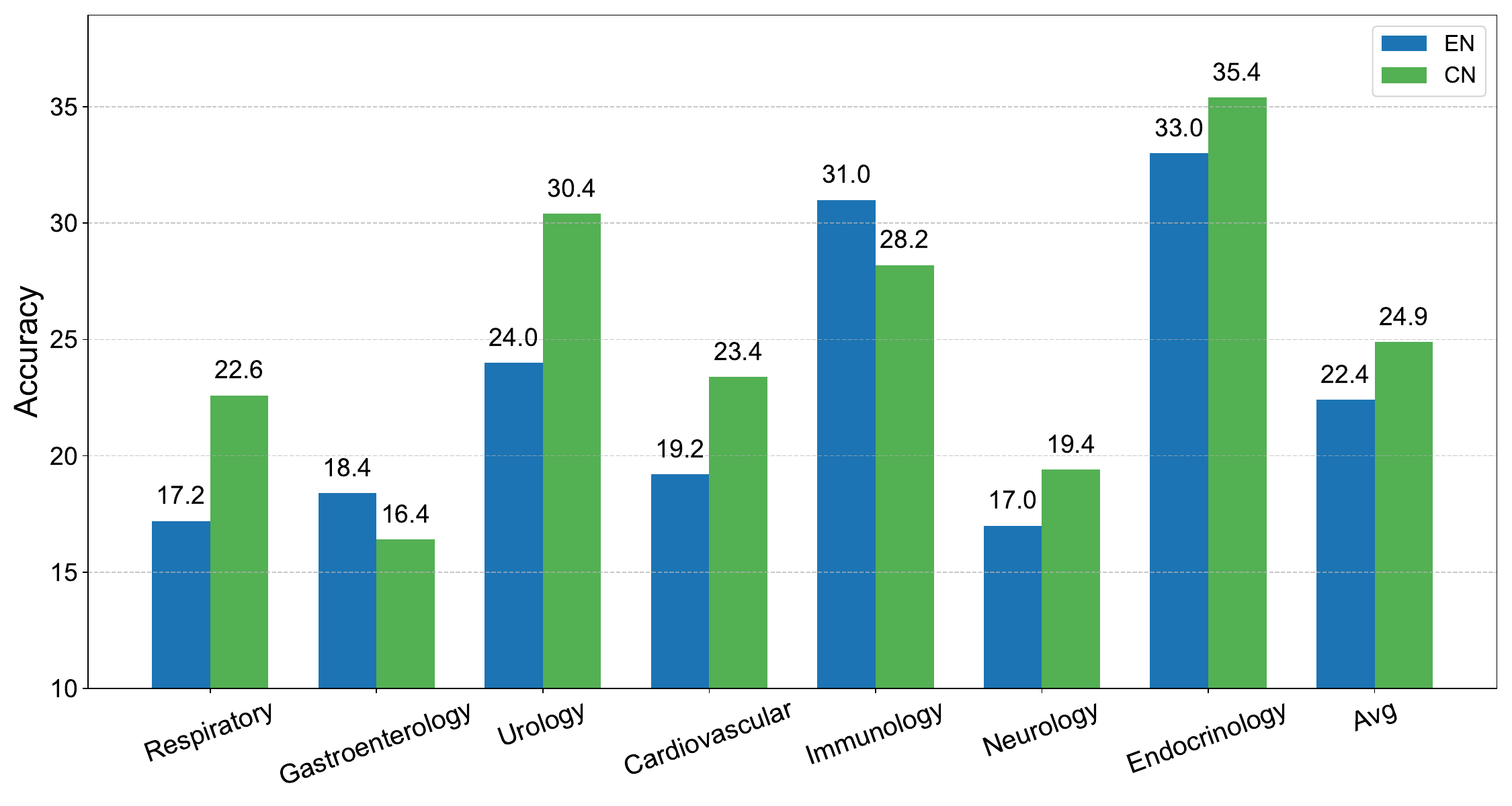} 
    \caption{Comparison of the Chinese and English versions of ClinicalGPT-R1} 
    \label{fig:3} 
\end{figure}

We also compared the training performance of the Qwen2.5-7B-Instruct model using Chinese and English datasets. For this comparison, we constructed Chinese and English training datasets based on the long reasoning data generated by GPT-4o-mini, and employed a two-stage reasoning approach for training. In addition, corresponding Chinese and English test sets were created. The experimental results indicate that the model trained on Chinese data outperforms the model trained on English data when evaluated on the Qwen2.5-7B-Instruct model Fig. ~\ref{fig:3}.

\begin{figure}[htbp]
    \centering
    \includegraphics[width=1\textwidth]{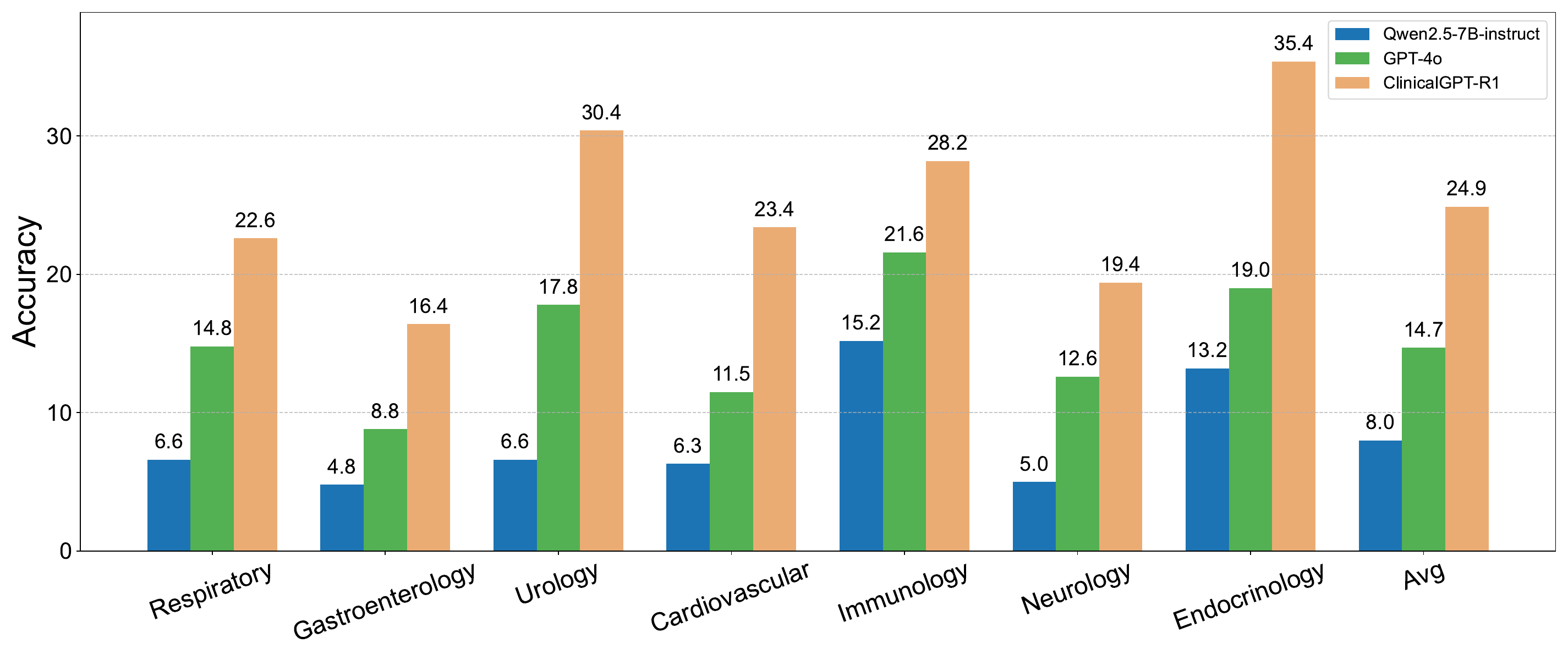} 
    \caption{Comparison of ClinicalGPT-R1 and baselines in Chinese} 
    \label{fig:4} 
\end{figure}

\begin{figure}[htbp]
    \centering
    \includegraphics[width=1\textwidth]{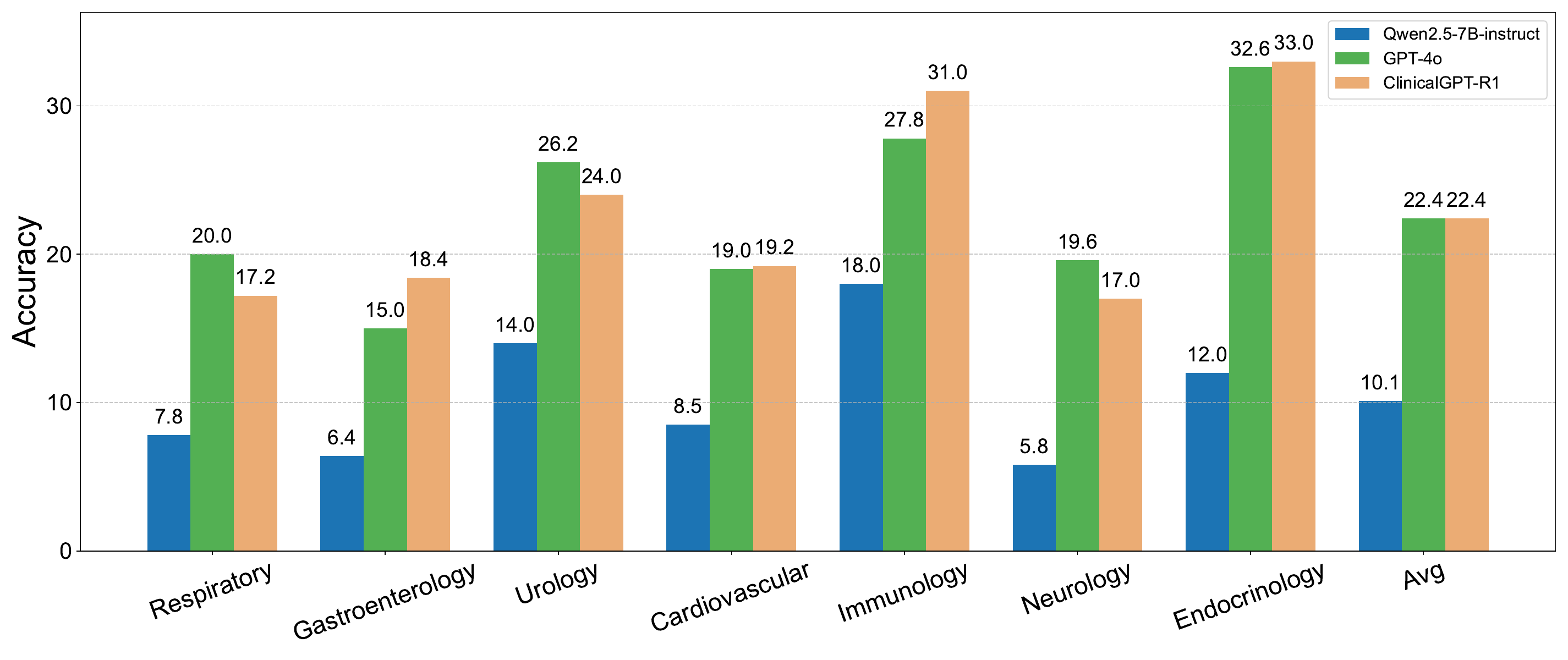} 
    \caption{Comparison between ClinicalGPT-R1 and baseline models in English} 
    \label{fig:5} 
\end{figure}

We also compared the diagnostic performance of ClinicalGPT-R1 with the base models Qwen2.5-7B-Instruct and GPT-4o in a multilingual environment. For the base models, reasoning was performed using the CoT prompting approach. In the Chinese version of the test, ClinicalGPT-R1 demonstrated significantly better diagnostic performance than GPT-4o and the Qwen2.5-7B-Instruct base model (as shown in Fig. ~\ref{fig:4}). In the English version of the test, ClinicalGPT-R1 performed comparable to GPT-4o and significantly outperformed the base model Qwen2.5-7B-Instruct (Fig. ~\ref{fig:5}), further validating the strong capabilities of the model in diagnostic reasoning.

Meanwhile, to test whether catastrophic forgetting occurred during the training process of ClinicalGPT-R1, we conducted tests on the MedQA. The results show that after ClinicalGPT-R1 developed strong medical reasoning capabilities, no catastrophic forgetting occurred.

\section{Conclusion}

In this study, we introduced ClinicalGPT-R1, a reasoning-enhanced LLM designed for clinical disease diagnosis. Recognizing that extended reasoning processes are critical in complex medical diagnostic scenarios, we systematically investigated various experimental setups to enhance the reasoning capabilities of LLMs. To address the complexity of medical reasoning, we constructed long-reasoning datasets via multiple synthesizers and explored diverse training strategies across languages. We also introduced MedBench-Hard, a challenging benchmark with stratified disease categories spanning seven medical specialties. Experimental results demonstrate that ClinicalGPT-R1 exhibits exceptional performance in disease diagnosis.

\section*{Acknowledgments}

This study was funded by the National Natural Science Foundation of China (grants 62272055, 62425112), New Cornerstone Science Foundation through the XPLORER PRIZE, and Young Elite Scientists Sponsorship Program by CAST (2021QNRC001).

\clearpage


\printbibliography


\end{document}